\documentclass[journal]{IEEEtran}

\usepackage{moreverb,url}
\usepackage{adjustbox}
\usepackage{algorithm}
\usepackage{algpseudocode}
\usepackage{subfig}
\usepackage{todonotes}
\usepackage{mwe}

\begin{document}
\title{Risk-Aware Planning by Confidence Estimation from Deep Learning-Based Perception}

\author{Maymoonah Toubeh and Pratap Tokekar%
\thanks{M. Toubeh is with the Department of Electrical \& Computer Engineering at Virginia Tech, Blacksburg, U.S.A. \tt{\small may93@vt.edu}.}
\thanks{P. Tokekar is with the Department of Computer Science at the University of Maryland at College Park, U.S.A. \tt{\small tokekar@umd.edu}.}}

\maketitle

\begin{abstract}
This work proposes the use of Bayesian approximations of uncertainty from deep learning in a robot planner, showing that this produces more cautious actions in safety-critical scenarios. The case study investigated is motivated by a setup where an aerial robot acts as a ``scout'' for a ground robot. This is useful when the below area is unknown or dangerous, with applications in space exploration, military, or search-and-rescue. Images taken from the aerial view are used to provide a less obstructed map to guide the navigation of the robot on the ground. Experiments are conducted using a deep learning semantic image segmentation, followed by a path planner based on the resulting cost map, to provide an empirical analysis of the proposed method. A comparison with similar approaches is presented to portray the usefulness of certain techniques, or variations within a technique, in similar experimental settings. The method is analyzed to assess the impact of variations in the uncertainty extraction, as well as the absence of an uncertainty metric, on the overall system with the use of a defined metric which measures surprise to the planner. The analysis is performed on multiple datasets, showing a similar trend of lower surprise when uncertainty information is incorporated in the planning, given threshold values of the hyperparameters in the uncertainty extraction have been met. We find that taking uncertainty into account leads to paths that could be 18\% less risky on an average.
\end{abstract}

\begin{IEEEkeywords}
risk-aware planning, uncertainty approximation, deep learning.
\end{IEEEkeywords}

\IEEEpeerreviewmaketitle

\section{Introduction}

 \begin{figure}[!tbp]
  \centering
  \subfloat[Handcrafted ground truth segmentation]{\includegraphics[width=1.65in]{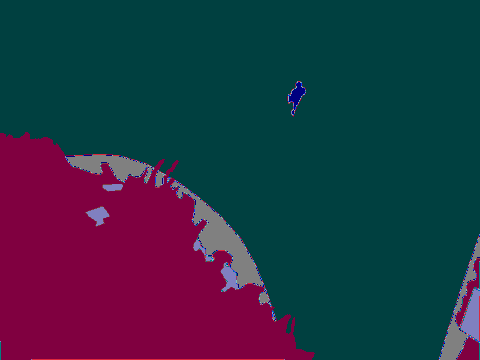}\label{fig:qual_gt}}
  \hfill
  \subfloat[Planning based on ground truth segmentation]{\includegraphics[width=1.65in]{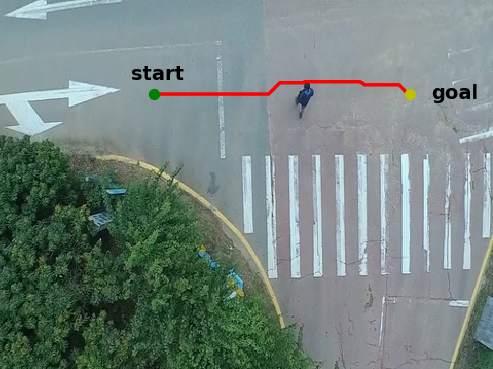}\label{fig:qual_gt_path}}
  \vfill
  \subfloat[Traditional deterministic segmentation]{\includegraphics[width=1.65in]{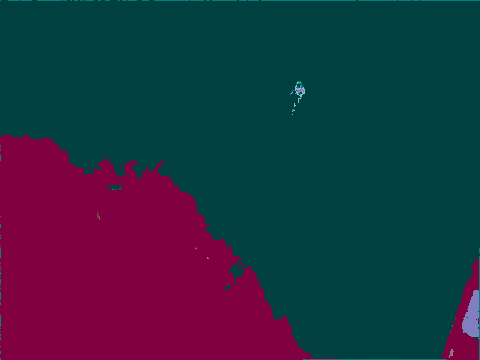}\label{fig:qual_deter_seg}}
  \hfill
  \subfloat[Planning based on DL model segmentation alone]{\includegraphics[width=1.65in]{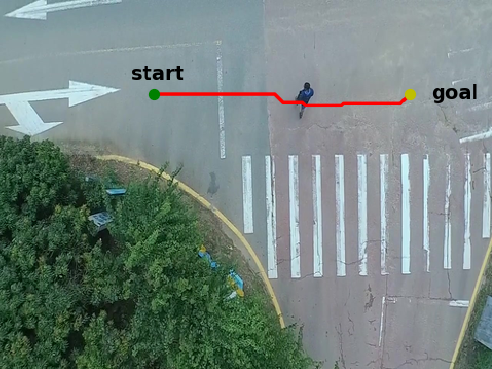}\label{fig:qual_deter_seg_path}}
  \vfill
    \subfloat[Uncertainty of DL model segmentation given by dropout]{\includegraphics[width=1.65in]{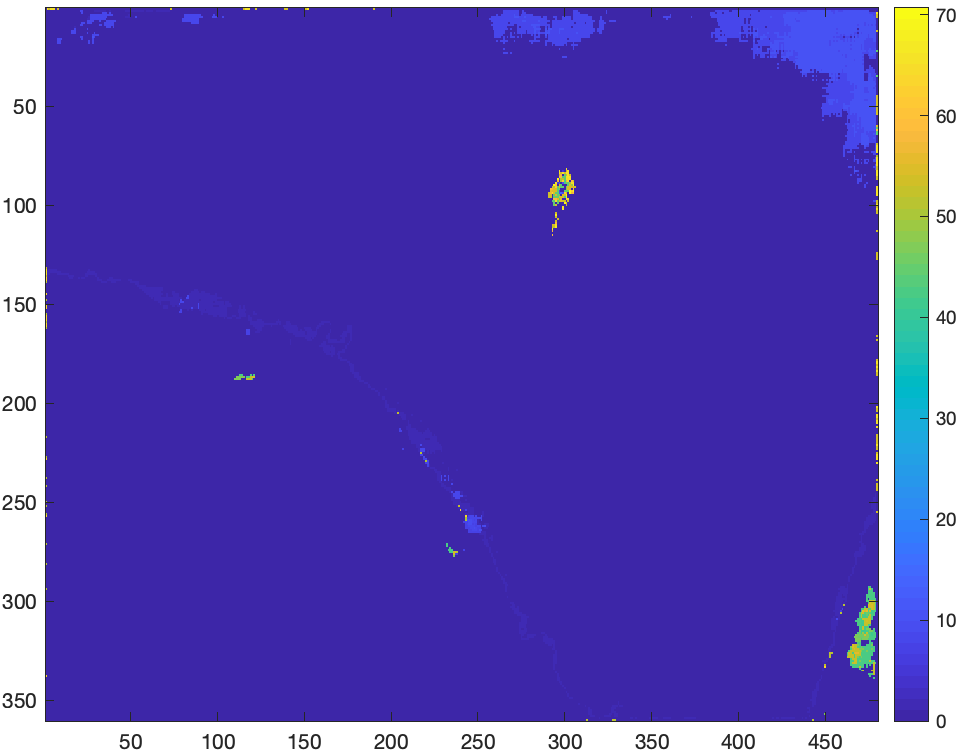}\label{fig:qual_unc_drops}}
  \hfill
  \subfloat[Planning based on DL model segmentation with uncertainty given by dropout]{\includegraphics[width=1.65in]{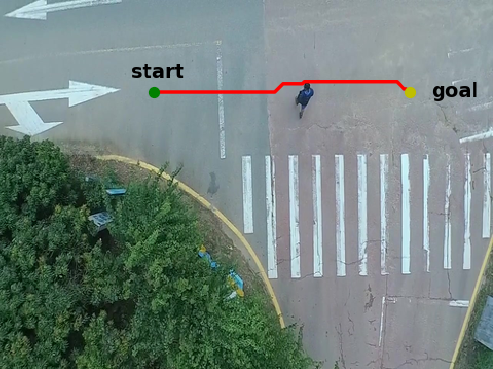}\label{fig:qual_unc_drops_path}}
  \vfill
    \subfloat[Uncertainty of DL model segmentation given by bootstraps]{\includegraphics[width=1.60in]{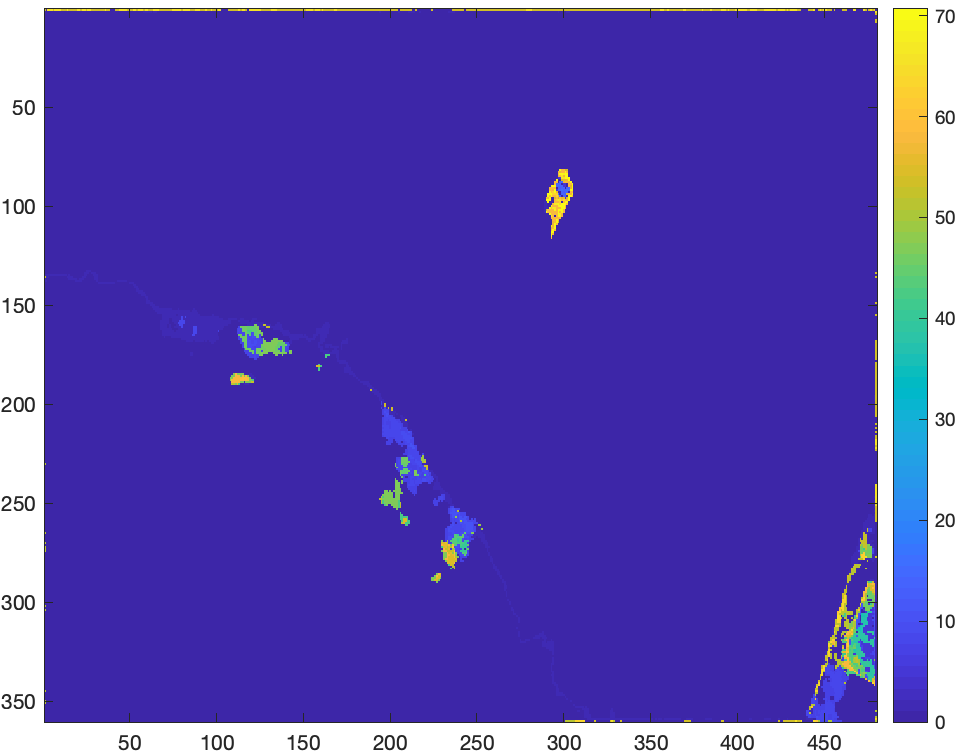}\label{fig:qual_unc_boots}}
  \hfill
  \subfloat[Planning based on DL model segmentation with uncertainty given by bootstaps]{\includegraphics[width=1.65in]{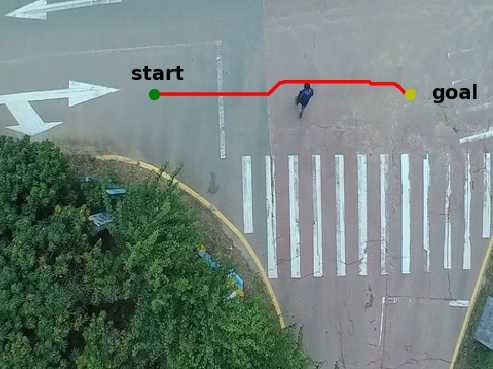}\label{fig:qual_unc_boots_path}}
  \caption{Qualitative results showing the path planned from start to goal given (a) handcrafted ground truth image segments, (b) deep learning model segmentation alone, and deep learning model segmentation with uncertainty given by (c) dropout and (d) bootstraps.}
  \label{fig:qual_results}
\end{figure}

\IEEEPARstart{R}ecent advances in deep learning (DL) algorithms, paired with significant improvements in hardware, have shown potential in many fields, including robotics. From enabling robot systems to navigate using high-dimensional image inputs, to allowing tractable trial-and-error robot learning both in simulation and reality; deep learning is everywhere. However, as promising as applications of DL to robot planning seem, the potential of the positive impact they may have on real-world scenarios is inevitably proportionate to their interpretability and applicability to imperfect environments.

For an example setup, this work utilizes a DL image segmentation model to generate a cost map used in an A* path planner. Figure~\ref{fig:qual_results} shows qualitative results given by the DL model and the subsequent planner. In this image taken from the Aeroscapes dataset~\cite{nigam2018ensemble}, the pedestrian near the top center is not sufficiently segmented by the DL model prediction shown in Figure~\ref{fig:qual_deter_seg}. Incorporating uncertainty associated with this prediction, given two extraction methods, before passing it to the planner produces a more reasonable and risk-aware path, as seen in Figures~\ref{fig:qual_unc_drops_path}~and~\ref{fig:qual_unc_boots_path}. Higher levels of uncertainty are visualized by hotter spots in Figure~\ref{fig:qual_unc_drops}~and~\ref{fig:qual_unc_boots}.

Deep learning is known for its data-driven rather than algorithmic learned representations~\cite{goodfellow2016dlbook}. A DL hierarchical structure can learn directly from data with little to no handcrafted features or learning variables~\cite{goodfellow2016dlbook}. However, this often comes at the expense of the interpretability of the learning outcomes. Deep neural networks can even misrepresent data outside the training distribution, giving predictions that are incorrect without providing a clear measure of certainty associated with the result~\cite{gal2017phd}. The outputs of a deep neural network are generally point estimates of the parameters and predictions present, so they do not provide a meaningful measure of correlation to the overall data distribution the network was trained on~\cite{gal2017phd}. For this reason, deep learning models are considered deterministic functions, often called ``black-boxes,'' unlike probabilistic models which inherently depict uncertainty information.

As a step towards risk-aware robotic systems that utilize the powers of DL, this work combines methods of approximating uncertainty in DL with robot planners that are partially reliant on an otherwise black-box approach. We utilize recently developed methods of uncertainty extraction from deep learning models~\cite{gal2017phd}. The extraction of uncertainty information, as opposed to the reliance on point estimates, is crucial in safety-critical applications, such as autonomous navigation in an urban setting. With the acquired uncertainty estimates, the system produces an explicable metric of its level of confidence and can therefore be altered to accommodate for risks in the environment.

In the robotics community, an emphasis has been placed on methods that work in a controlled experimental setup, but more recently risk-aware methods aim to ensure that these methods are also safe in the real-world. As a relatively new approach in robotics, techniques have been adapted from the fields of statistics and machine learning. Common statistical methods of accommodating risk include altering the optimization criterion so that it becomes risk-sensitive~\cite{garcia2015saferl}. Although modern DL models are usually considered black-boxes due to their mathematical nature, recent work has initiated theoretically grounded understandings of them. Such works investigate the integration of deep learning techniques with information theoretical and statistical approaches for the purpose of calculating model uncertainties, where uncertainty refers to an estimation of the model's lack of confidence in its predictions~\cite{gal2017phd}. It is these practical methods of quantifying risk associated with DL models that are utilized in the proposed planning systems of this paper. The uncertainty we focus on is epistemic, which is arguably more relevant in safety critical scenarios~\cite{kendall2017uncertainties}. 

The main contributions of this work are as follows:
\begin{itemize}
	\item We present a risk-aware framework for planning systems which already use deep learning for perception.
    \item We define a new risk-aware cost function as a means of interpreting uncertainty of a deep learning model's prediction in a meaningful way to a subsequent robot planner.
    \item We propose a new metric, termed \emph{surprise factor}, to provide a means of comparing what the robot planner expects with reality.
    \item We carry out an empirical analysis with the purpose of portraying the impact of changing the uncertainty extraction method, or hyperparameters within the same method, on the overall planning system. The surprise factor and runtime are observed, showing that the tuning of certain hyperparameters may be more important than others, especially in a resource constrained setting like robotics.
    \item The proposed risk-aware method is shown to produce more consistent and safe results, where the conventional risk-neutral method proves to be dangerous and possibly catastrophic, given that threshold values of the hyperparameters in the uncertainty extraction are met.
\end{itemize}
This paper expands on our preliminary work~\cite{toubeh2018risk}, with more rigorous testing of our proposed framework using more data, more parametric combinations, and different uncertainty extraction methods.

\section{Related Work}
Several prior works exist which extract uncertainty information from a deep learning model, mainly Bayesian neural networks, ensemble methods, and methods that utilize stochastic regularization techniques~\cite{gal2017phd}. Our work seeks to extend the most suitable of these approaches to robot planning.

\subsection{Bayesian Neural Networks}

Some of the earliest attempts to bind the reasoning of probabilistic models, such as Gaussian processes, with deep learning (DL) is seen in Bayesian neural networks (BNN)~\cite{kononenko1989bayesian,gal2017phd}. Unlike the DL models used in modern practice which do not depict uncertainty, a BNN produces an output that is a probability distribution over its predictions. Probability distributions are placed over the weights of a BNN, making it an approximation of a Gaussian process as the number of weights tends to infinity. Uncertainty can be extracted as a statistical measure, such as variance or entropy, over the output distribution of a BNN in order to capture how confident the model is with its prediction. However, BNN require a larger number of parameters to be trained, with less practical training methods available for them. A recent example of a Bayesian neural network utilizes a relatively small model and trains by minimizing alpha-divergences~\cite{depeweg2016bnn}.

\subsection{Ensemble Methods}

Since the practicality of Bayesian neural networks is questionable, approximations of these structures have arisen, one of which is ensemble methods. A recently developed ensemble method is referred to as the bootstrapped neural network, where several deep learning (DL) models with the same architecture are trained on subsets of the larger dataset sampled with replacement~\cite{kendall2017uncertainties}. The underlying concept behind this method is that the different bootstrap models will agree in high density areas and disagree in low density areas of the complete dataset. The outputs of the separate models combine to form a probability density function from which uncertainty of a prediction can be measured. This method is theoretically sound;however, it is not ideal for applications that are faced with time and resource constraints like robotics, as we show in our comparisons.

Ensemble approaches have been applied to several reinforcement learning problems, where the quantified uncertainty is utilized during learning to balance exploration and exploitation. One work proposes using bootstrapped Q-learning, demonstrating that using uncertainty to direct data collection produces faster learning that is also less expensive, tedious, or likely to lead to physical damage for a real robot~\cite{kalweit2017uncertainty}. Bootstrapping is one of the two uncertainty extraction methods analysed and compared in this work.

\subsection{Stochastic Regularization Methods}

Most recently, the use of stochastic regularization methods common in deep learning (DL) has been shown to also approximate Bayesian neural network models without changing the ready structures being used or their training process~\cite{gal2017phd}. Regularization is used as a means to avoid overfitting of a DL model to its training set, so that it generalizes to data that is similar enough but not exactly the same. This ensures the learning model has not simply memorized the training data, but has actually learned something meaningful about the data that will translate to a slightly different setting. At a high level, stochastic regularization techniques work by introducing randomness in the training process to increase the robustness of the model to noise. Dropout is one such popular method that is inspired by the probabilistic interpretations of deep learning models that consider activation nonlinearity a cumulative distribution function~\cite{goodfellow2016dlbook}. In its traditional use, dropout is only activated during training, in which case the weights of a deep learning model are randomly multiplied by zero or one in a certain predefined proportion. In this approach, at test time, weight averaging by the percentage of dropout applied during training is performed on the final trained model, which then leads to point estimate results. 

In order to form a distribution over the outputs of a model trained using a stochastic regularization technique such as dropout, the regularization is activated at test time, producing stochastic estimates with multiple passes of the same input through the model~\cite{gal2017phd}. The multiple stochastic passes are then averaged to form a mean estimate, as opposed to a point estimate, and uncertainty can be extracted given statistical or information theoretical metrics over the distribution. 

Stochastic regularization techniques have been previously adapted for uncertainty extraction in several problems, including reinforcement learning, semantic segmentation, camera relocalization, and robotic collision avoidance, among many others. One work applies dropout to a pre-existing model-based reinforcement learning algorithm in order to better quantify uncertainty over longer periods of learning~\cite{gal2016improving}. In another work, the uncertainty from dropout is used in semantic segmentation for improved learning and test time estimation~\cite{kendall2015bayesian}. In a similar work, the same approximation approach is applied to assist in camera relocalization for landmark detection in a SLAM problem~\cite{kendall2016modelling}. The uncertainty estimate is used to approximate the localization error with no additional hand-crafted parameterizations. In yet another work, the authors combine bootstrapped neural networks with stochastic regularization methods to avoid catastrophic or harsh collisions during robot training for collision avoidance~\cite{kahn2017collision}. They show that their method effectively minimizes dangerous collisions during training, while also showing comparable performance to baselines without explicit account for uncertainty. Unlike prior applications, our work provides a thorough analysis of the effect of variations in the hyperparameters involved in the uncertainty extraction process to the overall performance of the robot planning system. In this work, we do not simply assume a benefit of one uncertainty approximation approach, or variation within the same approach, over another, but instead provide an empirical comparison.

Newer work attempts to replace the multiple stochastic passes involved with sampling from stochastic regularization methods with a separately trained network with a sole purpose of uncertainty approximation. One of these instances uses a Mixture Density Network (MDN), where a single pass saves time in comparison to multiple in a robotics setting~\cite{choi2018uncertainty}. However, this comes at the expense of training a separate MDN for the task of uncertainty extraction. Our comparisons focus on methods of uncertainty approximation that do not involve a separate uncertainty model, but those that exploit the properties of deep learning models already trained. Previous research also suggests that as long as the number of stochastic passes used for sampling does not exceed the model's batch size, there is little to no negative impact on runtime~\cite{kendall2017uncertainties}. We show this is true in our empirical analyses.

\section{Approach}

\begin{table*}
\begin{center}
\small
\begin{adjustbox}{width=\textwidth}
\begin{tabular}{ c c c c c c c c c c c c c c }
    \hline
    \textbf{Softmax} & 1 & 2 & 3 & 4 & 5 & 6 & 7 & 8 & 9 & 10 & 11 & 12\\
	\textbf{Class} & Background & Person & Bike & Car & Drone & Boat & Animal & Obstacle & Construction & Vegetation & Road & Sky\\
    \textbf{Cost} & 20 & 140 & 130 & 90 & 7 & 80 & 120 & 100 & 110 & 5 & 1 & 150\\
    \textbf{Color} & \subfloat{\includegraphics[width=0.2in]{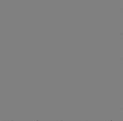}} & \subfloat{\includegraphics[width=0.2in]{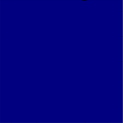}} & \subfloat{\includegraphics[width=0.2in]{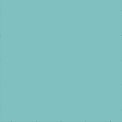}} & \subfloat{\includegraphics[width=0.2in]{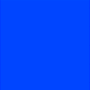}} & \subfloat{\includegraphics[width=0.2in]{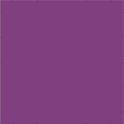}} & \subfloat{\includegraphics[width=0.2in]{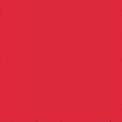}} & \subfloat{\includegraphics[width=0.2in]{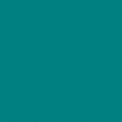}} & \subfloat{\includegraphics[width=0.2in]{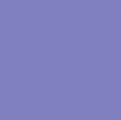}} & \subfloat{\includegraphics[width=0.2in]{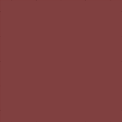}} & \subfloat{\includegraphics[width=0.2in]{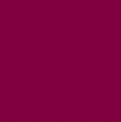}} & \subfloat{\includegraphics[width=0.2in]{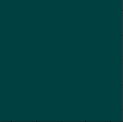}} & \subfloat{\includegraphics[width=0.2in]{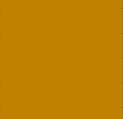}}\\
    \hline
\end{tabular}
\end{adjustbox}
\caption{The fixed costs assigned to each segmentation class from the Aeroscapes dataset to be used in A* search. These costs are hand-designed to generate qualitative examples that demonstrate the utility of the proposed approach. In the real world, classes that are not navigable (e.g. animal, car, bike) will be assigned infinite cost.\label{tab:costcolor}}
\end{center}
\end{table*}

\begin{figure}[!tbp]
  \centering
  \subfloat[Risk-neutral perception and planning]{\includegraphics[width=3.4in]{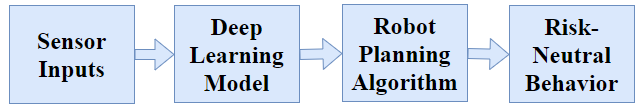}\label{fig:perceptneutral}}
  \vfill
  \subfloat[Risk-aware perception and planning]{\includegraphics[width=3.4in]{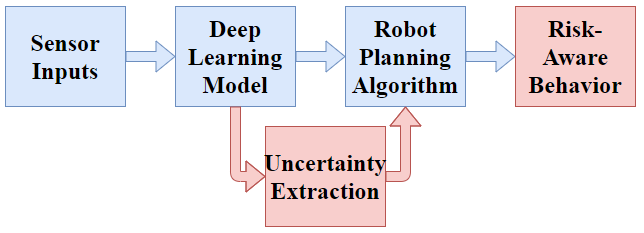}\label{fig:perceptrisk}}
  \caption{The overall flow of information in a robot planner that relies in part on a DL model in (a) the risk-neutral and (b) the risk-aware case.\label{fig:perceptloop}}
\end{figure}

The problem setup is inspired by a previous work~\cite{christie2017radiation} that uses the overhead imagery provided by an unmanned aerial vehicle (UAV) as input to an image segmentation algorithm, which is then used to assist the navigation of an unmanned ground vehicle (UGV). The UAV acts as a ``scout'' by flying ahead of the UGV. The overhead orthorectified imagery is then classified (into categories such as ``road'', ``person'', ``car'', etc.). Each category is assigned a cost (given in Table~\ref{tab:costcolor}) which is used to  determine a path for the ground robot to follow. The class cost is then augmented with the extracted uncertainty information provided by the deep learning model for the proposed risk-aware planning framework. Note that the fixed costs in Table~\ref{tab:costcolor} are different from those assigned in our previous work~\cite{toubeh2018risk}. After more thorough experimentation, it was found that assigning costs with a wider range giving more clear variation in scale for ``risky" classes provides a better basis for the risk-aware planner.

\subsection{Risk-Aware Cost}
Figure~\ref{fig:perceptloop} shows a high-level schematic of the proposed risk-aware approach by contrasting it with the traditional, risk-neural approach. Unlike the previous works, here, in addition to performing the semantic segmentation of the image, the uncertainty in the segmentation costs is also extracted. The measure of uncertainty is then used to manipulate the navigation away from low confidence regions. The navigation portion of the robot planner in this case is not a DL model, but a classical method, A* search. A cost function is mapped onto each semantic class (see Table~\ref{tab:costcolor}). Therefore, the uncertainty in the segmentation corresponds to uncertainties in the cost, for which A* search is sufficient. If the uncertainties in the transition function are to be considered instead, a Markov Decision Process (MDP) would be more useful. Considering uncertainty is contrast to trusting the outputs of the DL portion of the system invariably, which could lead to catastrophic outcomes if a point estimate outlier is produced in the case the input is considered out-of-distribution to the training data. In the proposed approach, an uncertainty metric can be used to calibrate the robot plan based on the level of confidence in the DL model predictions.

The results of the segmentation given by any DL model cannot guarantee complete accuracy in all settings. Variations in lighting, angle, or objects present in an image can contribute to inaccurate predictions. There will always be a prediction when a DL model is involved, as the model will force an estimate even when it does not make sense to. A good measure to account for this risk associated with DL outputs being used in the robot planner is to evaluate the certainty associated with the DL result. One practical method is using dropout, which is already being used as a regularizer during training. 

In the risk-neutral case, the pixel classification is taken as is from the DL model and assigned a cost accordingly. For the proposed risk-aware method, the cost is evaluated by adding the uncertainty value, multiplied by some factor, to the risk-neutral cost assignment. Specifically, the cost of pixel $p$ is given by,
\begin{equation}
C(p) = L(p) + \lambda \hat{V}(L(p)),
\label{eq:cost}
\end{equation}
where $L(p)$ is the cost associated with the semantic class that is predicted for $p$ (given in Table~\ref{tab:costcolor}) and $\hat{V}(p)$ is the uncertainty value extracted by dropout. In practice, $\hat{V}(p)$ does not need to be variance; it can be another statistical measure of uncertainty taken over the distribution provided by the stochastic passes. $\lambda$ is a weighting parameter. The risk-neutral case corresponds to $\lambda=0$ and the risk-aware case corresponds to $\lambda>0$.

We use the standard deviation over the label cost as the measure of uncertainty in our experiments. We found that standard deviation scales better than variance with the fixed cost function assigned in Table~\ref{tab:costcolor}. The scaling of the uncertainty metric would be less handcrafted if the model is calibrated~\cite{guo2017calibration}. However, this would require the training of an additional scaling parameter, not usually already applied in modern DL.

\subsection{Confidence Estimation}

\begin{algorithm}
\caption{Confidence Estimation using Dropout}\label{alg:uncextdrops}
\textbf{Input:} image $I$\\
\textbf{Output:} cost of average prediction for each pixel $L(p)$, average standard deviation for each pixel cost $\hat{V}(L(p))$\\
\begin{algorithmic}[1]
\State $p \gets$ pixels in image $I$
\For{$t$ = 1 to number of stochastic passes}
\State $O(p,c,t) \gets$ softmax output of stochastic pass $t$
\State $L(p,c,t) \gets$ costs associated with $O(p,c,t)$
\EndFor
\State $\hat{O}(p,c) \gets$ average $O(p,c,t)$ over stochastic passes
\State $Ind(p) \gets$ argmax of softmax in $\hat{O}(p,c)$
\State $L(p) \gets$ cost associated with $Ind(p)$
\For{$c$ = 1 to number of classes}
\State $V(L(p,c)) \gets$ standard deviation in $L(p,c,t)$ over stochastic passes for class $c$ 
\EndFor
\State $\hat{V}(L(p)) \gets$ average $V(L(p,c))$ over classes
\end{algorithmic}
\end{algorithm}

\begin{algorithm}
\caption{Confidence Estimation using Bootstraps}\label{alg:uncextboots}
\textbf{Input:} image $I$\\
\textbf{Output:} cost of average prediction for each pixel $L(p)$, average standard deviation for each pixel cost $\hat{V}(L(p))$\\
\begin{algorithmic}[1]
\State $p \gets$ pixels in image $I$
\For{$t$ = 1 to number of bootstraps}
\State $O(p,c,t) \gets$ softmax output of bootstrap model $t$
\State $L(p,c,t) \gets$ costs associated with $O(p,c,t)$
\EndFor
\State $\hat{O}(p,c) \gets$ average $O(p,c,t)$ over bootstraps
\State $Ind(p) \gets$ argmax of softmax in $\hat{O}(p,c)$
\State $L(p) \gets$ cost associated with $Ind(p)$
\For{$c$ = 1 to number of classes}
\State $V(L(p,c)) \gets$ standard deviation in $L(p,c,t)$ over bootstraps for class $c$ 
\EndFor
\State $\hat{V}(L(p)) \gets$ average $V(L(p,c))$ over classes
\end{algorithmic}
\end{algorithm}

Two methods of uncertainty extraction are compared in this work. Variations possible with each method are also compared and analyzed empirically to produce the experimental results of the next section. First, the overview of each uncertainty extraction technique is provided here.

Algorithm~\ref{alg:uncextdrops} shows a breakdown of uncertainty extraction using a stochastic regularization method to generate stochastic samples of the posterior. First, the stochastic outputs are generated for a number of stochastic passes, giving a softmax value for each pixel each time. For each pass, Bernoulli dropout is activated on the trained network, effectively multiplying random neuron weights by zero or one in a set proportion. The softmax outputs $O$ are averaged over all stochastic passes. The maximum value of this average softmax is taken as the output class prediction $Ind$, then the costs associated with the labels $Ind$ are saved in $L$ as in Table~\ref{tab:costcolor}. Standard deviation is computed over the stochastic passes for each output class costs, then the average of $V$ over all classes produces a single value for each pixel. The value $\hat{V}$ is considered the uncertainty in the cost associated with the pixel's prediction.

In Algorithm~\ref{alg:uncextboots}, $t$ represents the index of the bootstrap model being used to produce a stochastic sample, whereas it represents the index of the stochastic pass given the stochastic regularization at test time in Algorithm~\ref{alg:uncextdrops}. 

The uncertainty value is computed as the average standard deviation across all segment class costs for a particular pixel. The higher the average standard deviation, the less confident the DL model is with its prediction. Therefore, it is intuitive to incorporate this information along with the original prediction when planning a path for navigation, especially in a safety-critical environment such as a road.

\subsection{Deep Learning Model}
To demonstrate the utility of an uncertainty metric associated with a DL model being used as part of a robot planning system, experiments involving this setup are portrayed to compare the risk-neutral and risk-averse cases qualitatively. A dataset of aerial images is used to train the DL model, and then the model is tested on a different dataset to illustrate the robustness of the method to anomalies outside the training set.

We use the Bayesian SegNet to perform semantic segmentation of every pixel in the input image~\cite{kendall2015bayesian}. The Bayesian SegNet is first trained for the segmentation task using the predefined encoder-decoder architecture, along with the pretrained weights of the VGG16 image classification network~\cite{kendall2015bayesian}. Since the model is already well adapted for image classification, less further training is needed for per-pixel segmentation, in comparison to starting with random model weighting. A batch size of 2 is used to fit an 8GB Nvidia GeForce GTX 1080 GPU. 

Since the motivation for our work is inspired by prior applications in UAV-UGV coordination, where an autonomous aerial vehicle is responsible for providing a less obstructed map for the UGV below it, we trained the Bayesian SegNet on a realistic dataset from the aerial viewpoint. For this, the Aeroscapes dataset of 3269 diverse aerial images taken with a fleet of drones is used for training~\cite{nigam2018ensemble}. Aeroscapes is prelabeled to consist of 12 ground truth classes that correspond to 12 separate softmax outputs in the trained model. The classes and their softmax correspondence, as well as the color legend used for visualizations, are shown in Table~\ref{tab:costcolor}. The dataset is first divided into training, testing, and validation sets, allotting 1963, 653, and 653 images for each, respectively. The test accuracy is computed for each 360x480 pixel image, resized to fit the Bayesian SegNet inputs. The value for each image is then averaged over the 653 image test set to give a final test accuracy of 80.04\%. 

\begin{figure}[!tbp]
  \centering
  \subfloat[Input aerial image from Kentland]{\includegraphics[width=1.65in]{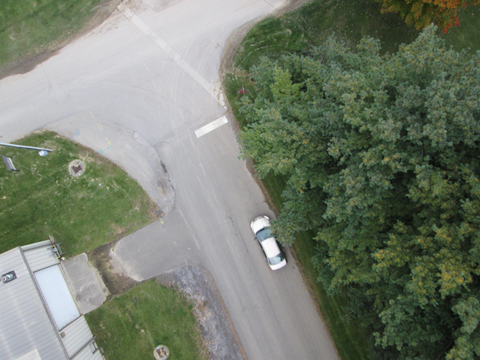}\label{fig:kentresults_img}}
  \hfill
  \subfloat[Traditional deterministic segmentation]{\includegraphics[width=1.65in]{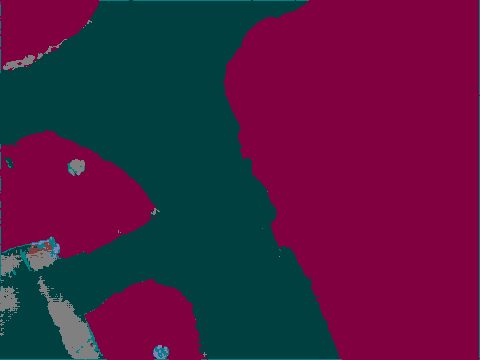}\label{fig:kentresults_deter_seg}}
  \vfill
  \subfloat[Dropout average segmentation]{\includegraphics[width=1.65in]{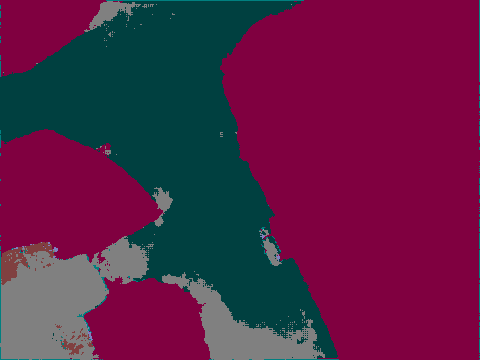}\label{fig:kentresults_drops_seg}}
  \hfill
  \subfloat[Uncertainty of DL model segmentation given by dropout]{\includegraphics[width=1.65in]{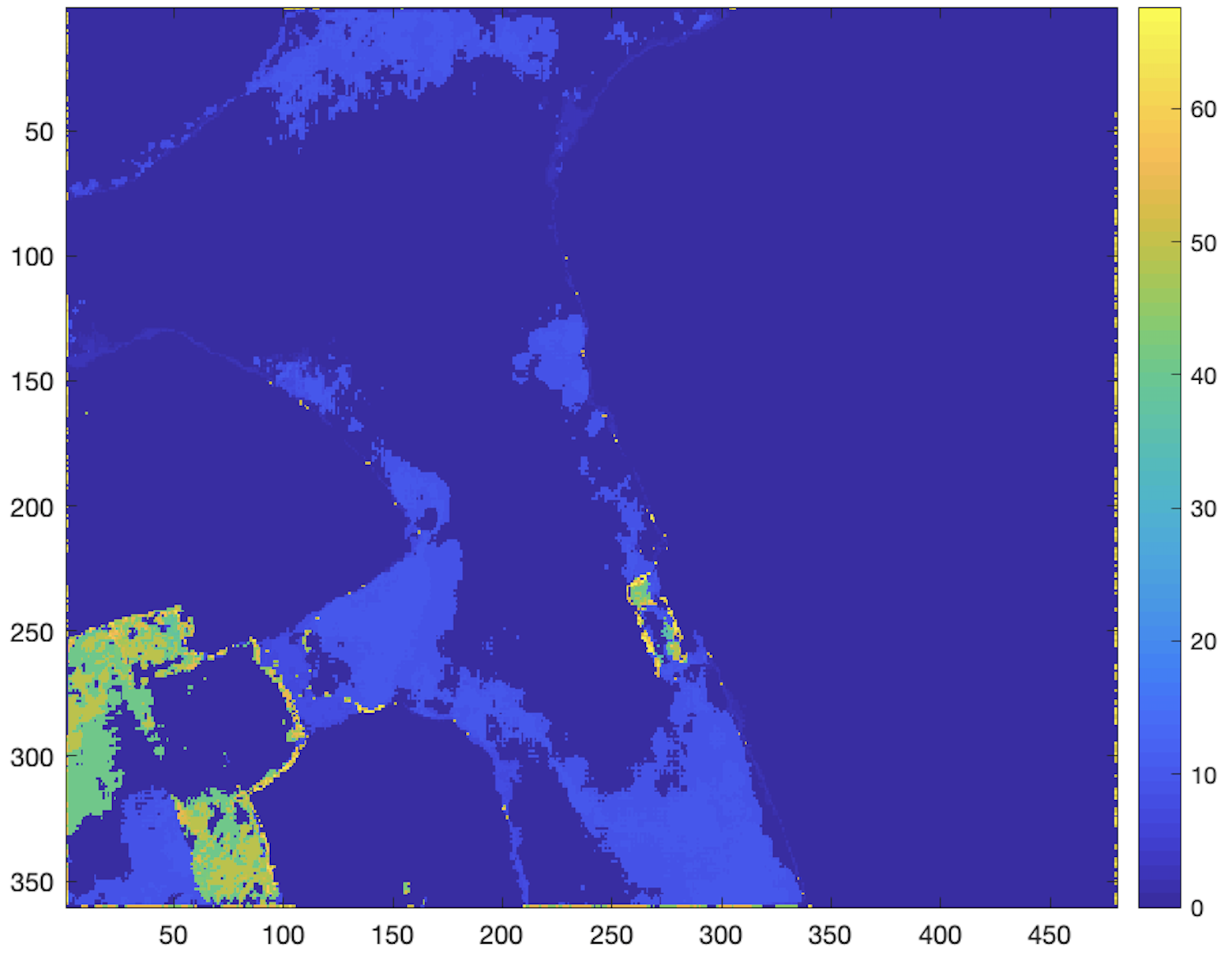}\label{fig:kentresults_drops_unc}}
  \vfill
  \subfloat[Bootstrap average segmentation]{\includegraphics[width=1.65in]{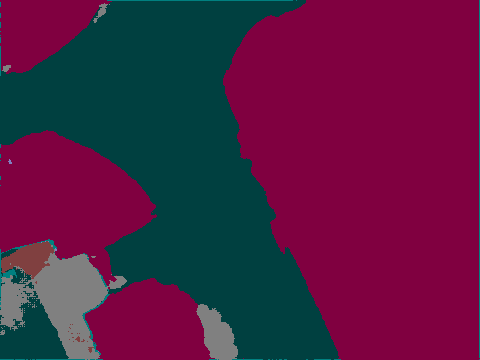}\label{fig:kentresults_boots_seg}}
  \hfill
  \subfloat[Uncertainty of DL model segmentation given by bootstraps]{\includegraphics[width=1.65in]{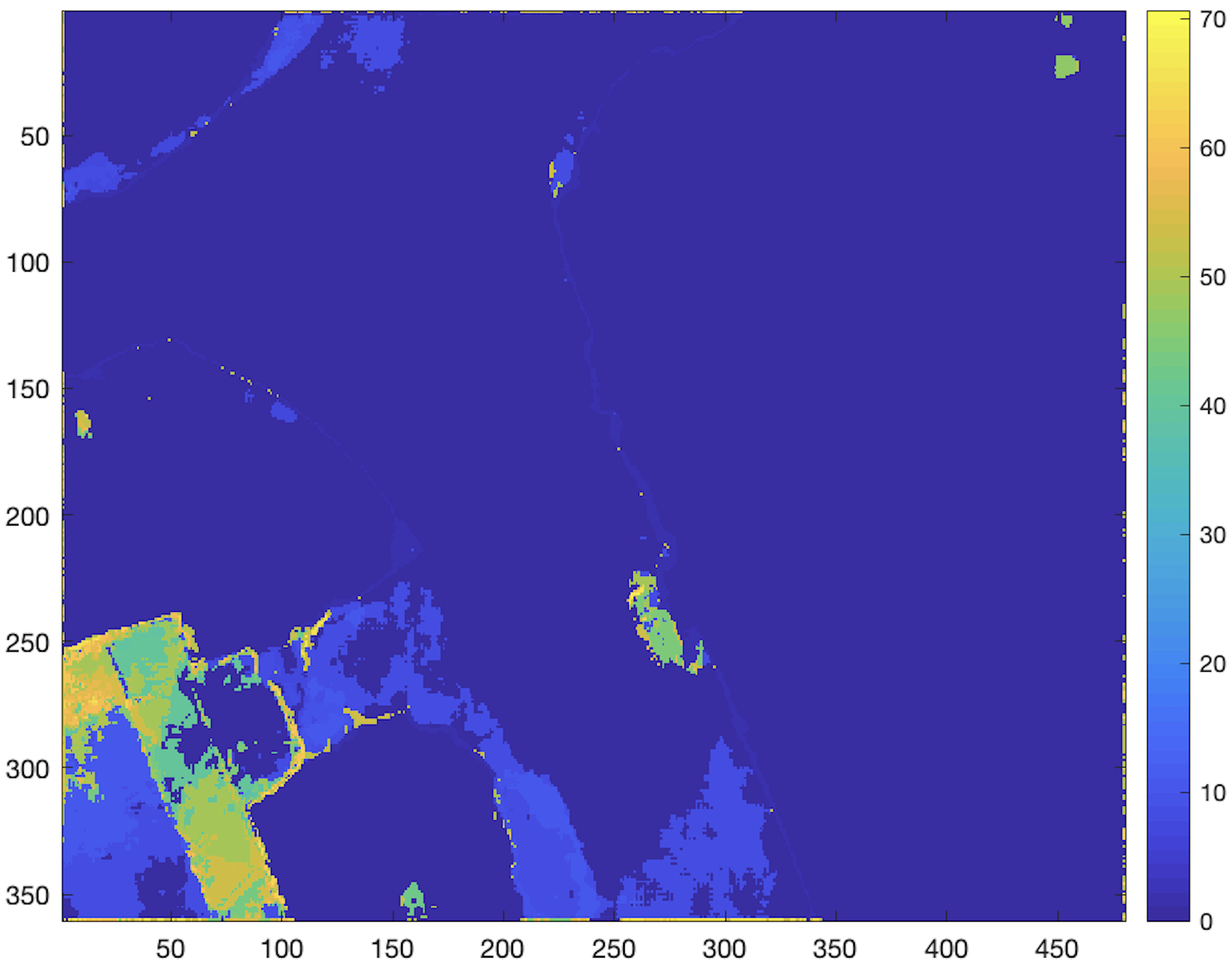}\label{fig:kentresults_boots_unc}}
  \caption{An example semantic segmentation produced by the Bayesian SegNet trained model on (a) an input image from the Kentland test set with (b) its traditional deterministic segmentation, (c) its average dropout segmentation, (d) the model uncertainty given using dropout, (e) its average bootstrap segmentation, and (f) the model uncertainty given using bootstraps.\label{fig:kentresults}}
\end{figure}

The trained segmentation model is also tested on a different aerial dataset taken from the authors of a previous UAV-UGV cooperative radiation detection work~\cite{christie2017radiation}. The dataset consists of 262 low flying UAV images taken at Kentland Farms at Virginia Tech, and it is labeled with only four of the ground truth labels in Aeroscapes: road, vegetation, construction, and car. A sample result from the Kentland dataset is shown in Figure~\ref{fig:kentresults}. Here, images from the Kentland dataset provide a means of testing the model on data for which it is not specifically trained, but similar performance is expected due to the similar view points and content. The building at the bottom left of the image seen in Figure~\ref{fig:kentresults_img} are misclassified by the network as a combination of multiple classes, corresponding to higher uncertainty. The car in the middle is consistently mislabeled ``background', also showing higher uncertainty. This shows that the uncertainty metric is not merely a function the average prediction, but it provides valuable information about the quality of the prediction in its own right. Errors are common, and expected, in a varying or real-world setting, but the uncertainty metric should provide a means of detecting such errors.

The dropout rate is set to 50\%, where half of the neurons in the six central encoder-decoder layers are set randomly to zero, consistent with the qualitative suggestion in~\cite{kendall2015bayesian}. Dropout is activated at training time as a regularizer and at test time to approximate the posterior over the output segmentation results. From the approximate posterior, the uncertainty metric is then extracted for use in the robot planner. The (not Bayesian) SegNet is also used for bootstrapping, by using the same model up to 10 times trained on up to 10 separate samples of the Aeroscapes training set taken with replacement.

\section{Results}

In order to perform a qualitative analysis of the risk-neutral and risk-aware methods, the \emph{surprise factor} is calculated to compare the expected path cost with the actual path cost by subtracting the two, then normalizing by the actual path cost for scaling. The path cost is found by summing up the cost associated with every pixel along the path. We use the predicted class of every pixel to determine the expected cost and the ground truth class of every pixel to determine the actual cost. If a path passes through pixels that the DL model classifies with low uncertainty, then we expect the predicted classes to be largely the same as the actual cost, thereby given a low surprise factor. On the other hand, if the predicted classes are wrong, then the surprise will be high.

Unlike Shannon entropy, the surprise factor defined here is proportionate to the value of a path (based on the cost function) and not how probable the path is. Shannon's theory of information addresses the accuracy with which a model depicts the data it is meant to, but not the semantic and subjective dimensions of the data~\cite{shannon1948mathematical,baldi2002computational}. In our work, we are concerned with the value associated with a prediction, and not only the accuracy of the prediction. A path cost prediction with a higher subjective value contributes more significantly to the surprise factor, but not its entropy. This formulation of the surprise factor is consistent with a risk-aware setting, where not all predictions have the same risk value. That being said, underestimation and overestimation of the path cost is treated symmetrically, given the absolute difference and division by the actual cost. If penalizing underestimates to a higher degree is desirable, the sign of the difference or division by the expected cost would address this.

\begin{table*}
\begin{center}
\small
\begin{tabular}{ c c c c c }
    \hline
	\textbf{Measure} & \textbf{Ground Truth} & \textbf{Deterministic} & \textbf{Dropout} & \textbf{Bootstraps}\\
    Expected Cost & 3200.158 & 1490.368 & 1345.695 & 1606.837 \\
    Actual Cost & 3200.158 & 7925.115 & 6910.737 & 7303.100 \\
    Surprise Factor & 0.000 & 0.591 & 0.484 & 0.427 \\
    \hline
\end{tabular}
\caption{The average surprise factor for 20 Aeroscapes images with diverse cost maps and 10 randomly chosen starting and goal positions each, given the average cost by ground truth, deterministic neural network, dropout, and bootstraps.\label{tab:aero_avg_costs}}
\end{center}
\end{table*}

Table~\ref{tab:aero_avg_costs} shows the trend for the average surprise factor calculated over 20 Aeroscapes test set images with 10 randomly chosen starting and goal positions. The surprise factor is calculated for the four different planning scenarios: using ground truth segmentation, using DL model segmentation alone, and using DL model segmentation while taking into account its prediction uncertainty using dropout or bootstrapping. When using DL segmentation alone, not accounting for model confidence increases the chance for a higher disparity between the expectation and reality. On the other hand, in the risk-aware approach, the expectation better matches the reality and leads to lower surprise.


\begin{figure}
\centering
\includegraphics[width=\columnwidth]{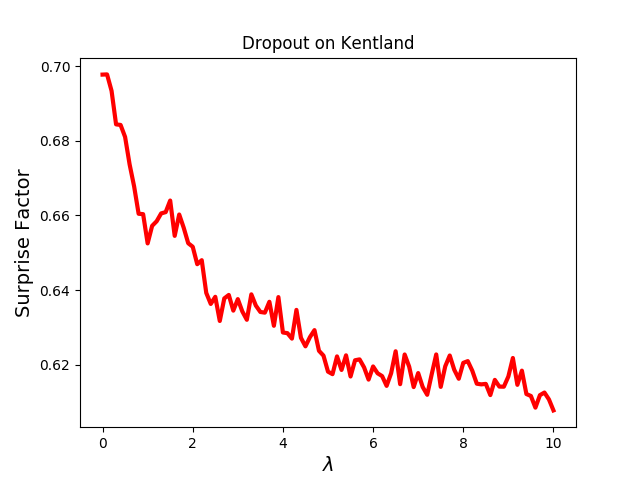}
\caption{Effect of varying $\lambda$ on the surprise factor, given uncertainty calculated using five stochastic passes with dropout, averaged over sixteen Kentland images with diverse cost maps given ten random start and goal position each.\label{fig:kent_lam_surp_d}}
\end{figure}

\begin{figure}
\centering
\includegraphics[width=\columnwidth]{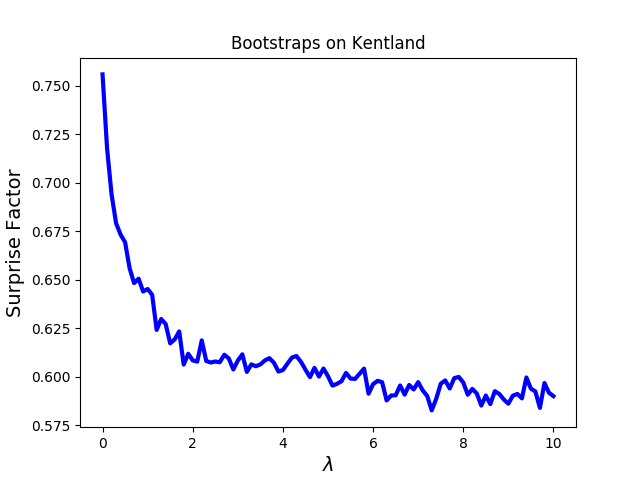}
\caption{Effect of varying $\lambda$ on the surprise factor, given uncertainty calculated using five bootstraps, averaged over 16 Kentland images with diverse cost maps given ten random start and goal position
each.\label{fig:kent_lam_surp_b}}
\end{figure}

\begin{figure}
\centering
\includegraphics[width=\columnwidth]{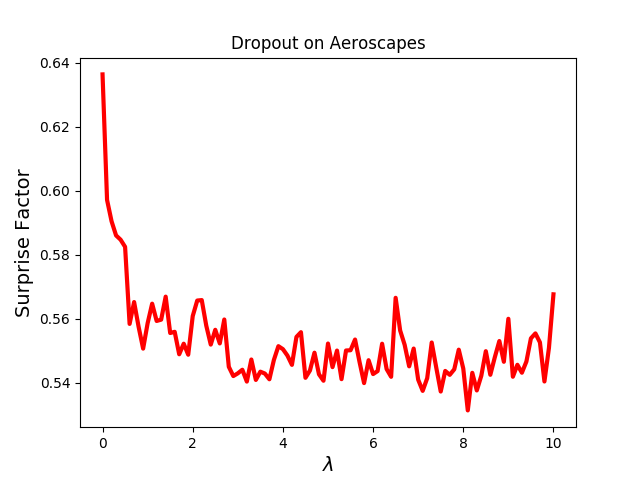}
\caption{Effect of varying $\lambda$ on the surprise factor, given uncertainty calculated using five stochastic passes with dropout, averaged over 20 Aeroscapes images with diverse cost maps given ten random start and goal position
each.\label{fig:aero_lam_surp_d}}
\end{figure}

\begin{figure}
\centering
\includegraphics[width=\columnwidth]{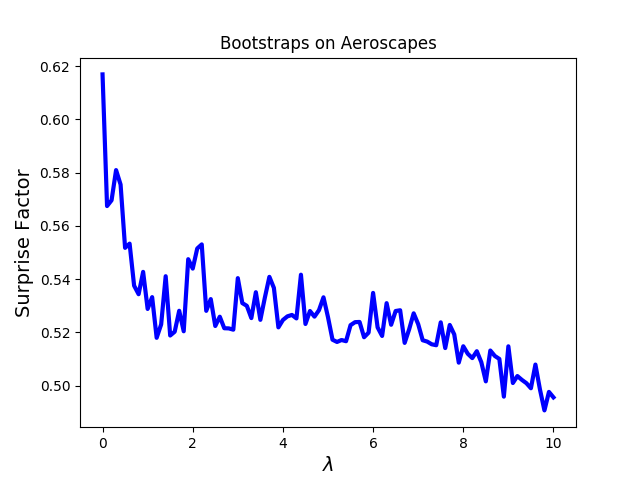}
\caption{Effect of varying $\lambda$ on the surprise factor, given uncertainty calculated using five bootstraps, averaged over 20 Aeroscapes images with diverse cost maps given ten random start and goal position
each.\label{fig:aero_lam_surp_b}}
\end{figure}

To find an appropriate value of $\lambda$, we perform an empirical analysis of the relationship between lambda and the defined surprise factor, as shown in Figures~\ref{fig:aero_lam_surp_d}~and~\ref{fig:aero_lam_surp_b}.  The number of dropout passes and bootstrap models is set consistently to five in this analysis. When $\lambda$ is small, the surprise factor is large. This is consistent with previous findings, since $\lambda=0$ corresponds to the risk-neutral case. As $\lambda$ increases, the surprise factor decreases finally converging to a fixed value. This is because, once $\lambda$ is sufficiently large, increasing $\lambda$ further does not change the path produced as output significantly (except for a few pixels). In fact, for very large $\lambda$, the path found will be the minimum uncertainty path since the second term in Equation~\ref{eq:cost} dominates the first term. Therefore, the surprise factor remains largely the same. Similar results are shown for a sample of the Kentland dataset in Figures~\ref{fig:kent_lam_surp_d}~and~\ref{fig:kent_lam_surp_b}.

\begin{table*}
\begin{center}
\small
\begin{tabular}{ c c c c c c c c c c }
    \hline
	\textbf{Bootstraps} & 2 & 3 & 4 & 5 & 6 & 7 & 8 & 9 & 10\\
	\textbf{Aeroscapes} & 0.436 & 0.437 & 0.440 & 0.418 & 0.435 & 0.405 &
 0.400 & 0.395 & 0.395\\
	\textbf{Kentland} & 1.018 & 1.124 & 0.912 & 0.892 & 0.964 & 0.878 & 0.850 & 0.866 & 0.889\\
    \hline
\end{tabular}
\caption{The average surprise factor for 20 Aeroscapes and 16 Kentland images with diverse cost maps and 10 randomly while varying the number of bootstrap models used.\label{tab:aero_kent_boots_surp}}
\end{center}
\end{table*}

\begin{table*}
\begin{center}
\small
\begin{tabular}{ c c c c c c }
    \hline
	\textbf{Dropout Passes} & 2 & 3 & 4 & 5 & 10\\
	\textbf{Aeroscapes} & 0.501 & 0.476 & 0.475 & 0.471 & 0.477\\
	\textbf{Kentland} & 0.776 & 0.704 & 0.761 & 0.771 & 0.79\\
    \hline
\end{tabular}
\caption{The average surprise factor for 20 Aeroscapes and 16 Kentland images with diverse cost maps and 10 randomly while varying the number of dropout passes used.\label{tab:aero_kent_drops_surp}}
\end{center}
\end{table*}

Tables~\ref{tab:aero_kent_drops_surp}~and~\ref{tab:aero_kent_boots_surp} show the trend of the surprise factor computed over varying numbers of dropout stochastic passes and bootstrap models, respectively. Since the value of $\lambda$ is fixed to eight for this analysis, as it is found to give lower surprise in the previous analysis, the change in surprise factor is relatively low. However, where there is change, the results are intuitive, showing lower surprise with higher sampling using more dropout passes or bootstrap models.

\begin{figure}
\centering
\includegraphics[width=\columnwidth]{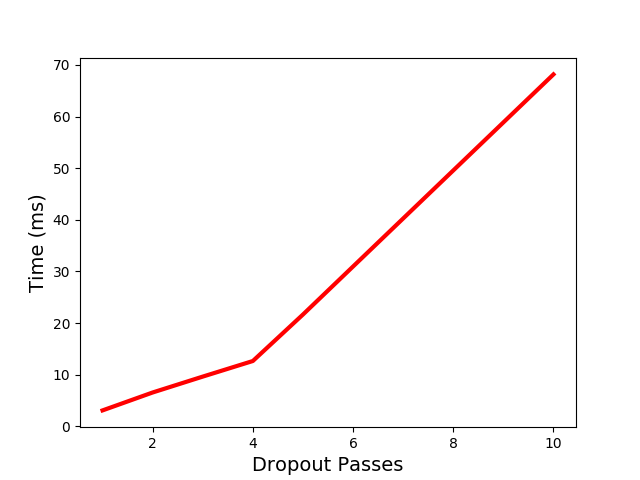}
\caption{Effect of varying the number of stochastic passes on the runtime (milliseconds) averaged over 16 Kentland images.\label{fig:kent_stoch_time_d}}
\end{figure}

\begin{figure}
\centering
\includegraphics[width=\columnwidth]{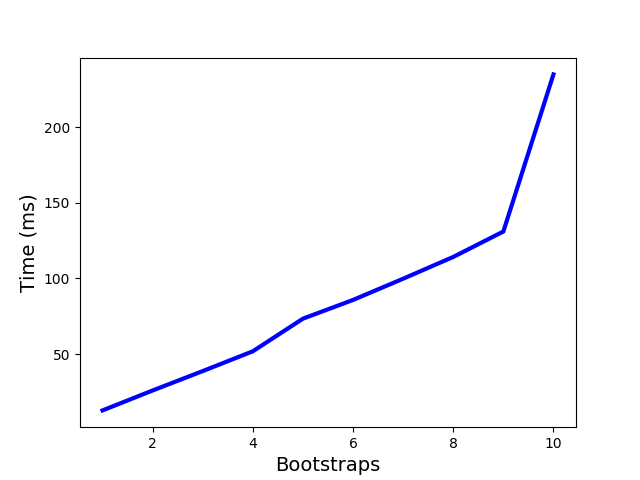}
\caption{Effect of varying the number of bootstraps on the runtime (milliseconds) averaged over 16 Kentland images.\label{fig:kent_boot_time_b}}
\end{figure}

The runtime for each method is also analyzed over the number of samples using dropout and bootstrapping. As seen in Figure~\ref{fig:kent_stoch_time_d}, while the number of dropout passes is less than five, the GPU's maximum parallel capacity, there is little change in runtime. However, once the number of passes is increased beyond five, they must run in sequence, increasing runtime significantly. Since the impact of using more stochastic passes on the surprise factor is low, saving runtime might be preferable to the designer of a resource constrained system. When using bootstrapping, as shown in Figure~\ref{fig:kent_boot_time_b}, runtime increases more quickly and is orders of magnitude higher than dropout. The bootstrap model parameters need to be read from memory individually. This is much more time consuming than applying dropout to the same model parameters, with little difference in the surprise factor. Bootstrapping might be more applicable offline, while dropout is more realistic for online applications. A more conservative risk-aware approach might combine the highest uncertainty value of either dropout or bootstrapping, if time can be spared. Time calculations are dataset independent because images are all resized to fit the neural network input. 

\section{Conclusions and Future Work}
This work proposes a risk-aware approach to robot planning that already involves deep learning. Risk is quantified by the model prediction uncertainty in the planning process. When deep learning is used for perception as a portion of the planning loop, an understanding of confidence in DL estimates is useful. Uncertainty is extracted directly from the DL models utilizing dropout and bootstrapping as practical methods. Promising results show that including uncertainty in a planner provides better predictability of actions, and even the avoidance of catastrophic actions in a safety-critical setting. Future work will involve providing probabilistic safety guarantees as in previous planners that also respect the uncertainty in the dynamics of a real robot and its environment~\cite{pairet2018uncertainty}. Another ongoing work is exploring the potential of the proposed framework as input to a risk-aware multi-robot task assignment problem~\cite{zhou2018approximation}.

\section*{Acknowledgment}
The material is based upon work supported by the National Science Foundation under Grant number 1566247 and the Office of Naval Research under Grant N000141812829.

\bibliographystyle{acm}
\bibliography{main.bib}

\end{document}